\documentclass[journal]{IEEEtran}
\usepackage{cite}
\usepackage{array}
\usepackage{amsmath,amssymb,amsfonts}
\usepackage{algorithmic}
\usepackage[linesnumbered,ruled,vlined]{algorithm2e}
\usepackage{graphicx}
\usepackage{textcomp}
\usepackage{xcolor}
\usepackage{enumerate}
\usepackage{booktabs}
\usepackage{diagbox}
\usepackage{array}
\usepackage{supertabular}
\usepackage{caption}
\usepackage{float} 
\usepackage{subcaption}
\usepackage{verbatim}
\usepackage{bm}
\usepackage{breqn}
\usepackage{stfloats}
\usepackage{amsthm}
\usepackage{hyperref}

\begin{document}

\title{Hybrid-Generative Diffusion Models for Attack-Oriented Twin Migration in Vehicular Metaverses}
	
\author{Yingkai Kang, Jinbo Wen,  Jiawen Kang*, Tao Zhang, Hongyang Du, Dusit Niyato, \textit{Fellow, IEEE}, \\ Rong Yu, and Shengli Xie, \textit{Fellow, IEEE}

\thanks{
        Y. Kang, J. Kang, R. Yu, and S. Xie are with the School of Automation, Guangdong University of Technology, Guangzhou 510006, China (e-mails: 3122000883@mail2.gdut.edu.cn; kavinkang@gdut.edu.cn; yurong@gdut.edu.cn; Shlxie@gdut.edu.cn).
        
        J. Wen is with the College of Computer Science and Technology, Nanjing University of Aeronautics and Astronautics, Nanjing 210016, China (e-mail: jinbo1608@163.com). 

        T. Zhang is with the School of Software Engineering, Beijing Jiaotong University, Beijing 100044, China (e-mail: taozh@bjtu.edu.cn).
        
        H. Du and D. Niyato are with the School of Computer Science and Engineering, Nanyang Technological University, Singapore (e-mails: hongyang001@e.ntu.edu.sg; dniyato@ntu.edu.sg). 

        \textit{Corresponding author: Jiawen Kang.}
	} 
}
	
\maketitle

\begin{abstract}
The vehicular metaverse is envisioned as a blended immersive domain that promises to bring revolutionary changes to the automotive industry. As a core component of vehicular metaverses, Vehicle Twins (VTs) are digital twins that cover the entire life cycle of vehicles, providing immersive virtual services for Vehicular Metaverse Users (VMUs). Vehicles with limited resources offload the computationally intensive tasks of constructing and updating VTs to edge servers and migrate VTs between these servers, ensuring seamless and immersive experiences for VMUs. However, the high mobility of vehicles, uneven deployment of edge servers, and potential security threats pose challenges to achieving efficient and reliable VT migrations. To address these issues, we propose a secure and reliable VT migration framework in vehicular metaverses. Specifically, we design a two-layer trust evaluation model to comprehensively evaluate the reputation value of edge servers in the network communication and interaction layers. Then, we model the VT migration problem as a partially observable Markov decision process and design a hybrid-Generative Diffusion Model (GDM) algorithm based on deep reinforcement learning to generate optimal migration decisions by taking hybrid actions (i.e., continuous actions and discrete actions). Numerical results demonstrate that the hybrid-GDM algorithm outperforms the baseline algorithms, showing strong adaptability in various settings and highlighting the potential of the hybrid-GDM algorithm for addressing various optimization issues in vehicular metaverses.
\end{abstract}

\IEEEpeerreviewmaketitle

\section{Introduction}

\IEEEPARstart{T}{he} vehicular metaverse is a virtual space that operates parallel to the physical world~\cite{35}. It seamlessly integrates the physical and virtual worlds through technologies such as eXtended Reality (XR), Artificial Intelligence (AI), and Digital Twin (DT) technology~\cite{lee2021all}, providing Vehicular Metaverse Users (VMUs) with an immersive experience, that is expected to bring revolutionary changes to the automotive industry~\cite{kang2024metaverses}. For example, the Augmented Reality (AR) head-up display solution uses AR technology to project important information such as navigation information, traffic signs, and driving routes directly onto the windshield, allowing the driver to obtain key information without diverting his or her gaze. This seamless information integration improves driving safety and significantly reduces the cognitive load of drivers. Vehicle Twin (VT) is a key technology in the vehicular metaverse~\cite{kang2024blockchain}, enabling real-time monitoring and management of vehicles by creating high-precision digital replicas that cover the entire vehicle life cycle~\cite{35}. VTs synchronize real-time vehicle status and traffic conditions from the physical space to the virtual space, enabling the vehicle to operate in a dynamic virtual environment~\cite{27}. Vehicles can provide VMUs with various metaverse services through VTs, such as allowing VMUs to experience virtual travel in the form of an avatar.

Constructing and updating VTs require substantial computing and storage resources. Due to the limitations in local computing resources of vehicles~\cite{khayyat2020advanced}, computation-intensive tasks for VT construction and update have to be offloaded to edge servers. However, maintaining service stability is challenging due to the high vehicular mobility and the uneven distribution of edge servers. Hence, VT migrations are conducted from the current edge servers to other edge servers~\cite{14}, which ensures seamless virtual experiences for VMUs. To reduce the load pressure on the current edge servers, vehicles adopt a pre-migration strategy to pre-migrate part of VT tasks~\cite{13}, thereby effectively utilizing computing resources, reducing data processing delays, and ensuring a seamless and high-quality user experience. Considering that the vehicular metaverse will integrate 6G technology to achieve space-air-ground-sea integrated networks~\cite{9806418}, satellites can be used to supplement the computing power of RoadSide Units (RSUs).

Unfortunately, there are security risks during the VT migration process. For example, attackers launch Distributed Denial of Service (DDoS) attacks on the edge servers to paralyze the VT migration or obtain the data uploaded by vehicles through the Co-resident attack~\cite{32,33}. To ensure that vehicles migrate VTs to reliable edge servers, the reputation value of edge servers needs to be evaluated through trust evaluation methods. Most trust evaluation methods are based on user reviews in DT migration scenarios, among which the subjective logic model is the most common~\cite{35,37}. Nevertheless, the traditional subjective logic model only quantifies the reputation value of edge servers according to the evaluation of VMUs, which makes the reputation evaluation of edge servers not objective enough. Besides, there may be malicious evaluators among VMUs, which may damage the reliability of reputation evaluation. 

To address the above challenges, we propose a secure and reliable VT migration framework in vehicular metaverses. Specifically, we design a two-layer trust evaluation model to compute the reputation values of edge servers. This model ensures the security of VT migrations while safeguarding the immersive experience for VMUs. Considering that the VT migration problem is NP-hard~\cite{13}, and traditional algorithms struggle to find optimal solutions within an acceptable time frame, we propose a hybrid-GDM algorithm based on Deep Reinforcement Learning (DRL), which utilizes a diffusion model to generate hybrid actions (i.e., continuous actions and discrete actions), thereby generating an optimal migration strategy for vehicles. The main contributions of this paper are summarized as follows:
\begin{itemize}
    \item To provide VMUs with immersive experiences, we propose a novel VT migration framework with high security and reliability. In the VT migration framework, vehicles select the appropriate and reliable edge servers (e.g., satellites and RSUs) to achieve efficient and reliable VT migrations, enabling real-time updates of VTs within the vehicular metaverse.
    \item Considering the potential security threats of edge servers, we propose a two-layer trust evaluation model. At the network communication layer, the reputation value of the edge server is evaluated based on its historical defense data and performance in detection tasks. In the interaction layer, the reputation value of an edge server is evaluated based on the interaction evaluation of the edge server by VMUs. By integrating the reputation values of the edge server at the network communication layer and interaction layer, the overall credibility of the edge server can be comprehensively quantified.
    \item To effectively identify the optimal VT migration strategy, we first model the VT migration decision problem as a Partially Observable Markov Decision Process (POMDP). Then, we propose the hybrid-GDM algorithm, which combines the diffusion model with advanced DRL algorithms to enhance the optimization potential. In previous work, GDMs can only generate continuous actions~\cite{23} or discrete actions~\cite{20}, while the hybrid-GDM algorithm is designed to generate hybrid actions to effectively optimize the migration strategy.
\end{itemize}

The rest of the paper is organized as follows. Section~\ref{s2} summarizes the related work. System overview is introduced in Section~\ref {s3}. The VT migration optimization problem formulation is described in Section~\ref{s4}. Section ~\ref{s5} describes a scheme for making VT migration decisions using the proposed hybrid-GDM algorithm. Section~\ref{s6} demonstrates the evaluation of multiple performances. Section~\ref{CONCLUSION} concludes this paper.

\section{Related Work}\label{s2}

\subsection{Vehicular Metaverse}\label{s2-1}
Neal Stephenson first depicted the concept of the metaverse in his science fiction novel \textit{Snow Crash} in 1992~\cite{27}. The metaverse is a digital world constructed by the integration of Internet and network technologies along with XR, aiming at simulating and enhancing interactions and experiences in real life~\cite{lee2021all}. The authors in~\cite{wang2022survey} discussed a novel distributed metaverse architecture and the foundational technologies of the metaverse, highlighting key issues in security and privacy protection. In~\cite{zhou2022vetaverse}, the authors defined Vetaverse as a future continuum between the vehicles industry and metaverses, envisioning it as the DTs of intelligent transportation systems. Most metaverse services are computationally intensive~\cite{cai2023joint}. In~\cite{feng2023resource}, the authors optimized the allocation of computing resources in vehicular metaverses through an intelligent scheduling algorithm, thereby improving the performance and user experience of AR applications. The authors in~\cite{27} proposed a contract model based on the migration task age to incentivize RSUs to provide bandwidth resources and ensure the continuity and service effectiveness of the VTs migration in the vehicular metaverse. The author in~\cite{13} proposed a framework for optimizing VTs migration using real-time trajectory prediction and multi-agent DRL, aimed at enhancing user immersion in the vehicular metaverse and reducing resource consumption. However, existing work rarely considers the security issues during VT migration in the vehicular metaverse.

\subsection{Digital Twin Migration}\label{s2-2}
Virtual machine migration refers to transferring a virtual machine between different physical hardware units, and has been studied in detail in~\cite{9}. For example, the authors in~\cite{11} introduced the dynamic virtual machine migration scheduling strategy, which significantly reduces computing service latency. Some recent studies have introduced the DT to optimize the migration process. DTs are the digital replica that covers the life cycle of their physical counterparts, i.e., Physical Twin, such as a physical object, process, or system~\cite{56}. In~\cite{12}, the authors introduced a wireless DT edge network model leveraging DRL, aiming to effectively reduce DT migration delay. The authors in~\cite{14} optimized the VT migration framework to address the high vehicular mobility and the unpredictable load on edge servers. The authors in~\cite{kang2024tiny} proposed a tiny machine learning-based Stackelberg game framework to achieve efficient DT migration in the emerging paradigm of the UAV metaverse through a tiny MADRL algorithm to address the challenges of UAV dynamic mobility and limited communication coverage of ground base stations. However, existing work does not consider the impact of potential security threats from edge servers on users during the VT migration.

\subsection{Generative Diffusion Models}\label{s2-3}
As an emerging GAI technology, GDMs are widely used in various fields such as image and video generation~\cite{57} and molecular design~\cite{58} due to their powerful content generation capabilities. GDMs achieved impressive results in the field of wireless communications. For example, the authors in~\cite{du2024enhancing} provided a comprehensive tutorial on applying GDMs to network optimization tasks, demonstrating the potential of GDMs to solve complex optimization problems inherent in networks. The authors in~\cite{69} used GDMs to generate optimal incentive mechanism solutions to reduce the overhead of generative mobile edge networks. In~\cite{20}, the authors used GDMs to generate optimal discrete decisions for the AIGC service provider selection problem. In~\cite{23}, the authors used GDMs to generate optimal contract designs to incentivize users to share semantic information. However, previous works were designed for discrete and continuous action spaces, respectively, which cannot be applied to hybrid action spaces.

\begin{figure*}[!htbp]
\centering
\includegraphics[width=0.9\textwidth]{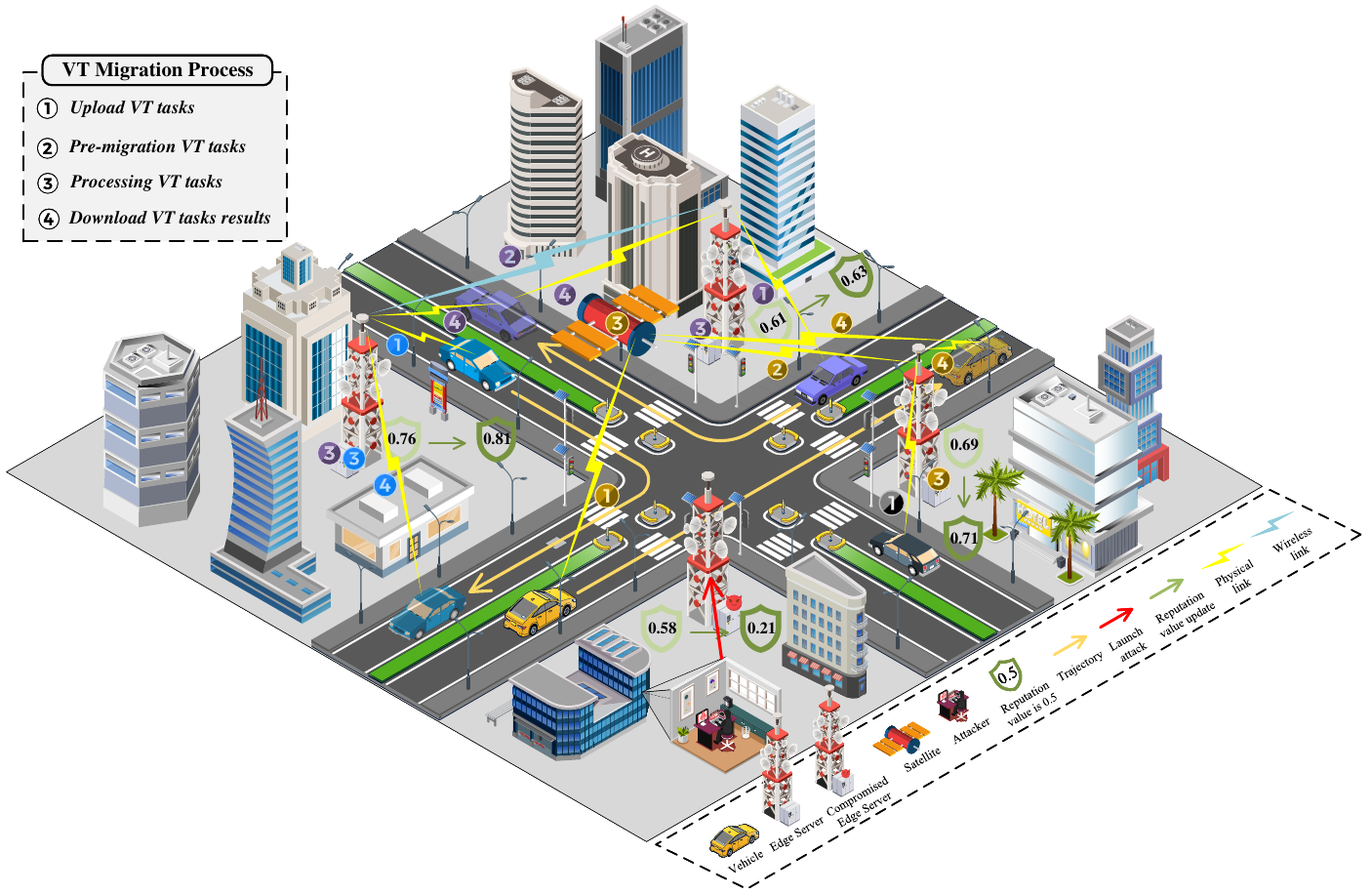}
\caption{Secure and reliable vehicle twin migration framework for vehicular metaverses.}
\label{fig1}
\end{figure*}

\section{Secure and Reliable Vehicle Twin Migration Framework}
\label{s3}

In this section, we introduce the proposed framework in vehicular metaverses, including the VT migration model, attack models, the two-layer trust evaluation model, and utility functions. Fig.~\ref{fig1} demonstrates that each vehicle will migrate the VT (e.g., avatar) to an edge server with a good reputation value for updates in each time slot, and some of VTs update tasks are pre-migrated to the subsequent edge servers with a good reputation for processing (i.e., the pre-migration process). After the edge servers complete the update of the VTs, they send the results back to the vehicles. In the vehicular metaverse, there are multiple vehicles and multiple edge servers, where the vehicle set is represented as $\mathcal{V}=\{1,\ldots,v\ldots, V\}$ and the edge server set is expressed as $\mathcal{S}=\{1,\ldots,s,\ldots, S\}$. Let $s_{p}$ represent the edge server where the vehicle pre-migration VT task arrives, $L_s^{max}$ represent the maximum load capacity of the edge server $s$, ${R}_{s}^{com}$ represent the communication coverage of the edge server $s$, and $c_{s}$ represent the computing capability of the edge server $s$. Time is divided into discrete time slots $\mathcal{T}=\{1,\ldots,t,\ldots, T\}$~\cite{39}.

\subsection{Migration Model}
\label{s3-1}

Initially, the delay from migrating VTs to an edge server ${s}$ (e.g., an RSU or a satellite) is calculated. Let $P_s=(x_s,y_s,z_s)$ represent the spatial coordinate of the edge server $s$, and $P_v(t)=(x_v(t),y_v(t),z_v(t))$ represent the spatial coordinate of vehicle $v$ at time slot $t$. Thus, the Euclidean distance between the edge server $s$ and the vehicle $v$ at time slot $t$ can be calculated as
\begin{equation}
\resizebox{\linewidth}{!}{$P_{v,s}(t)=\sqrt{[x_s-x_v(t)]^2+[y_s-y_v(t)]^2+[z_s-z_v(t)]^2}$}.
\label{eq1}
\end{equation}

The propagation model of the link between the edge server and the vehicle consists of Line-of-Sight (LoS) and Non-Line-of-Sight (NLoS) channels~\cite{yuan2022uav}. As vehicles move, their wireless communication channels with edge servers will change. Considering the uniformity of the wireless transmission channels~\cite{13}, both uplink and downlink, the path loss channel gain~\cite{14} for the link between the vehicle $v$ and the edge server $s$ can be calculated as
\begin{equation}
h_{v,s}(t)
=\begin{cases}A^{rsu}[\frac{c}{4\pi fP_{v,s}(t)}]^2,\text{$s$ is RSU},\\(A^{sat}_{los}+A^{sat}_{nlos})[\frac{c}{4\pi fP_{v,s}(t)}]^2,\text{$s$ is satellite}.
\end{cases}
\label{eq2}
\end{equation}

Here, $A^{rsu}$, $A^{sat}_{los}$, and $A^{sat}_{nlos}$ represent the gain for the RSU channel, LoS channel, and NLoS channel, respectively, $c$ is the speed of light, and $f$ is the carrier frequency. In the wireless communication scenario, the transmission delay generated when the vehicle $v$ sends the VT task to the edge server $s$ is determined by the uplink transmission rate, which can be calculated as
\begin{equation}
R_{v,s}^{up}(t)=B^{up}_s\log_2\left[1+\frac{p_vh_{v,s}(t)}{\sigma_s^2}\right],
\label{eq3}
\end{equation}
where $B^{up}_s$ is the uplink bandwidth, $p_v$ is the transmission power of the vehicle $v$, and ${\sigma_s^2}$ is additive Gaussian white noise at the vehicle $v$. When VMUs require vehicular services, the vehicle will offload the VT task $D_{v}^{up}$ to the edge server currently serving. Thus, the uplink transmission delay caused by migrating the VT to the edge server $s$ can be calculated as $T_{v,s}^{up}(t)=\frac{D_{\nu}^{up}(t)}{R_{\nu,s}^{up}(t)}$.

\begin{table}[!t]
\centering
\caption{Key Notations in the Paper}
\label{tab2}
\begin{tabular}{>{\centering\arraybackslash}p{2.5cm}>{\raggedright\arraybackslash}p{5cm}} \toprule
 \textbf{Parameters}& \textbf{Descriptions} \\ \midrule
$\mathcal{S}, {S}$& Set and number of edge servers, respectively\\
$\mathcal{V}, V$& Set and number of vehicles, respectively\\
$\mathcal{T}, T$& Set and  length of time slots, respectively\\
 $L_s(t)$&Load of edge server $s$ at time slot $t$\\
 $c_s$&Computing capability of edge server $s$\\
$P_v(t)$, $P_s$& Spatial coordinates of vehicle $v$ at time slot $t$ and edge server $s$, respectively\\
$B^{up}_s, B^{down}_s$& Uplink and downlink bandwidth of edge server $s$, respectively\\
 $B_{s,s_p}$&Physical link
bandwidth between edge server $s$ and $s_p$\\
 $R^{up}_{v,s}, R^{down}_{v,s}$&The rate of uplink and downlink between vehicle $v$ and edge server $s$, respectively\\
 $D^{up}_{v}$&VT task size uploaded by vehicle $v$\\
 $D^{task}_{v}$& VT task size to be processed by the edge server\\
 $D^{mig}_{v}$& Pre-migration task size\\
 $K^{pre}_{v}$&Pre-migration task proportion\\
 $e_v$&GPU cycles per unit data for vehicle $v$\\
 $T^{up}_{v,s}$&Upload latency from vehicle $v$ to edge server $s$\\
 $T^{que}_{s}$&Queuing delay of edge server $s$\\
 $T^{mig}_{s,s_p}$&Migration latency between edge server $s$ and $s_p$\\
 $T^{pro}_{v,s}, T^{pro}_{v,s_p}$&The processing delay from the edge server $s$ and $s_p$ receiving a task request to completing the task, respectively\\
 $T^{down}_{v}$&Downlink delay for vehicle $v$ to download the VT task result\\ 
 $T^{sum}_v$&Total latency of VT task of vehicle $v$\\
 $D^{tot}_s, D^{abr}_s$&Total and abnormal data sizes of the detection task, respectively\\
 $R^{tot}_{s}, R^{suc}_{s}$&Number of total requests and successful responses, respectively\\
 $P^{data}_{s}, P^{ser}_{s}$&Data security and service performance of edge server $s$,  respectively\\
 $\beta_{s}$&Defense performance of edge server $s$\\
 $E^{tot}_s$&Total number of evaluations received by edge server $s$\\
 $Rep_s^{Net}, Rep_s^{Int}$&Reputation values of the network communication and interaction layers of edge server $s$, respectively\\
 $Rep_s$&Comprehensive reputation value of edge server $s$\\
 $Rep_{s}^{new}$&Updated reputation value of edge server $s$\\\bottomrule
\end{tabular}
\end{table}

After the VT task is offloaded to the edge server, it must wait for the edge server to process it. This queuing time depends on the current load $L_s(t)$ and computing capability $C_{s}$ of the edge server $s$. Hence, the queuing delay can be expressed as $T_{s}^{que}(t)=\frac{L_{s}(t)}{c_{s}}$. To ensure a seamless experience for VMUs, vehicle $v$ can pre-migrate some part of VT tasks from the current edge server $s$ to the edge server $s_p$, and the VT tasks are pre-migrated from the current server $s$ to the pre-migration server $s_p$ through entity links between edge servers. Consequently, the edge server $s_p$ can process VT tasks simultaneously with the current server $s$, thereby optimizing resource allocation and utilization. We define the proportion of pre-migrated tasks as $K_{v}^{pre}(t)\in[0,1]$. The physical link bandwidth between these servers is defined as $B_{s,s_{p}}$. Thus, the delay caused by the pre-migration part of the VT task can be calculated as
\begin{equation}
T_{s,s_p}^{mig}(t)=\frac{D_{v}^{mig}(t)}{B_{s,s_p}},
\label{eq4}
\end{equation}
where $D_{v}^{mig}(t)=K_{v}^{pre}(t)D_{v}^{task}(t)$ is the total size of pre-migration tasks and $D_{v}^{task}(t)$ is the task size that needs to be processed by the edge server $s$. After other tasks are processed, the edge server handles the VT tasks of vehicle $v$. The delay in processing from the current edge server $s$ receiving a task request to completing the tasks is calculated as
\begin{equation}
T_{v,s}^{pro}(t)=T_{s}^{que}(t)+\frac{e_v(D_v^{task}(t)-D_{v}^{mig}(t))}{c_s},
\label{eq5}
\end{equation}
where $e_v$ denotes the number of GPU cycles needed to process a unit of data from vehicle $v$. For the edge server $s_p$, the processing delay of the pre-migration server $s_p$ can also be calculated as
\begin{equation}
T_{v,s_p}^{pro}(t)=T_{s,s_p}^{mig}(t)+T_{v,s_p}^{que}(t)+\frac{e_v D_{v}^{mig}(t)}{c_{s_p}}.
\label{eq6}
\end{equation}

Since the current edge server $s$ and the pre-migration server $s_p$ process VT tasks simultaneously, the delay of this process can be expressed as
\begin{equation}
T_{v}^{pro}(t)=\max\{T_{v,s}^{pro}(t),T_{v,s_{p}}^{pro}(t)\}.
\label{eq7}
\end{equation}

In addition to offloading VT tasks, vehicles may receive results (e.g., AR navigation routes) from different edge servers processing VT tasks. Specifically, vehicle $v$ will offload the VT task to the edge server $s$ and pre-migrate part of the VT task to the edge server $s_p$ based on the migration decision. When edge servers $s$ and $s_p$ finish processing the VT task, the results will be returned to the vehicle $v$. Similar to the uplink rate, the downlink rate of the edge server $s$ returning the processed results to the vehicle $v$ is calculated as $R_{v,s}^{down}(t)=B^{down}_s\log_2[1+\frac{p_v h_{v,s}(t)}{\sigma_s^2}]$. Thus, the delay experienced by vehicle $v$ when downloading the VT task results can be calculated as
\begin{equation}
T_{v}^{down}(t)=\frac{D_{v}^{task}(t)-D_{v}^{mig}(t)}{R_{v,s}^{down}(t)}+\frac{D_{v}^{mig}(t)}{R_{v,s_p}^{down}(t)}.
\label{eq8}
\end{equation}

In summary, for the vehicular metaverse, when vehicle $v$ performs a VT migration task at time slot $t$, it first sends a VT migration request to edge server $s$ based on the migration decision $K_v(t)$, which is discussed in detail in Section~\ref{s5-1}, resulting in uplink delay $T_{v,s}^{up}(t)$. The edge server $s$ then migrates some of the VT task to the edge server $s_p$ for processing. These two servers process the task in parallel, resulting in processing delay $T_{v}^{pro}(t)$. After the task is completed, the result is sent back to the vehicle $v$, and this process will generate a downlink delay $T_{v}^{down}(t)$. Therefore, the total delay for the VT migration can be calculated as
\begin{equation}
T_{v}^{sum}(t)=T_{v,s}^{up}(t)+T_{v}^{pro}(t)+T_{v}^{down}(t).
\label{eq9}
\end{equation}

During the VT migration, there are security risks in virtual-real interaction and large amounts of data transmission between edge devices~\cite{luo2023privacy}. Attackers may use specific strategies to attack edge servers and degrade migration efficiency. To better evaluate the reliability of edge servers, we discuss the main attacks faced in migration scenarios in Section~\ref{s3-2}.

\begin{figure*}[!ht]
\centering
\includegraphics[width=0.9\textwidth]{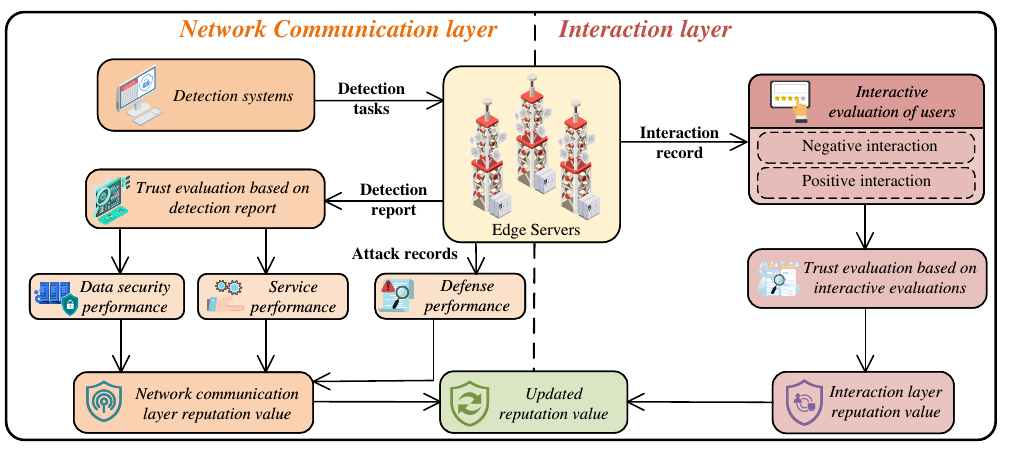}
\caption{Two-layer trust evaluation model.}
\label{fig2}
\end{figure*}

\subsection{Attack Model}
\label{s3-2}

We consider two attacks against edge servers: DDoS attacks~\cite{he2021game} and Co-resident attacks~\cite{45}. In the DDoS attack, attackers flood the target server with a large amount of malicious traffic, making it unable to handle legitimate requests. Since VT migrations rely on the physical network composed of edge servers, VT migrations may be forced to stop when these edge servers are attacked. In the co-resident attack, the attacker hides in an edge server, remains dormant as long as possible, and then leaks information stored on the edge server through side-channel attacks. Attackers take advantage of restricted resources to launch a variety of attacks. To maximize the effect, attackers employ strategies as follows:

\begin{itemize}
  \item [1)] \textbf{Direct DDoS attacks:}
  Malicious traffic is sent straight to a target edge server, flooding the link and pushing the current load $L_{s}(t)$ of edge server $s$ to its maximum capacity $L_s^{max}(t)$. When performing trust evaluation on the edge server $s$, resource exhaustion and service unavailability cause a significant increase in abnormal data and unresponsive requests. Consequently, there are significant reductions in the data security and service performance of the edge server $s$. Additionally, the edge server $s$ will receive negative evaluations due to service unavailability at this time.
  \item [2)] \textbf{Indirect DDoS attacks:}
  Attackers do not directly target the edge server but indirectly affect it by attacking nearby edge servers (e.g., crossfire attack~\cite{hyder2021towards}). Overloading nearby edge servers that share critical links prevents them from effectively processing transmitted data, increasing network congestion, and resulting in an increase in load on the target edge server and a decrease in overall responsiveness. The impact of this attack on the neighboring edge servers is similar to suffering a direct DDoS attack.
  \item [3)] \textbf{Co-resident attacks:}
  Attackers attempt to coexist their malicious script with the target edge server, using side-channel attacks to steal information from the target server and affect its performance. At this time, the edge server $s$ generates unexpected load and computing overhead due to the existence of malicious scripts. When performing detection tasks on edge servers, the presence of malicious scripts can slightly increase the volume of abnormal data and the incidence of abnormal response failures.
\end{itemize}

To prevent vehicles from migrating VTs to compromised edge servers that suffer the above attacks, we designed a two-layer trust evaluation model to effectively quantify the reliability of edge servers, thereby achieving efficient VT migrations and ensuring a seamless virtual experience for VMUs. The evaluation model is discussed in Section~\ref{s3-3}.

\subsection{Two-layer Trust Evaluation Model}
\label{s3-3}

Figure~\ref{fig1} shows that the reliability of edge servers is evaluated when migrating VTs from vehicles to these servers, which is critical to ensuring service quality and network security. This directly affects the efficiency of data processing and VT migration. To effectively measure the performance and credibility of edge servers, we design a two-layer trust evaluation model, as shown in Fig.~\ref{fig2}. This trust evaluation model comprehensively evaluates the reputation value of the edge server by considering its historical performance, task completion rate, service quality, and user feedback.

The first step of trust evaluation occurs at the network communication layer, evaluating the data security and service performance during edge server communication. After an edge server is attacked, the transmitted data might be intercepted, altered, or damaged~\cite{wang2020topology}. The capability of the edge server to maintain data security and integrity is measured by the ratio of non-anomalous data identified during task detections issued by detection systems (e.g., Snort and Suricata)~\cite{sharma2021retracted}. Specifically, the ratio of anomalous data is identified by following way~\cite{70}: (1) Verify the integrity of the data format and encapsulation; (2) Verify the consistency of data content with the parameters specified in the packet header; (3) Verify the consistency of multiple data contents at the same time. A higher ratio indicates that the edge server has more robust capabilities in protecting data from corruption or alteration, ensuring reliability during VT migrations. Therefore, the data security of the edge server $s$ can be calculated as~\cite{70}
\begin{equation}
P_s^{data}=\frac{D_s^{tot}-D_s^{abr}}{D_s^{tot}},
\label{eq10}
\end{equation}
where $D_{s}^{tot}$ is the total data size of the detection task and $D_{s}^{abr}$ is the abnormal data size of the detection task. After being attacked, the edge server may become overloaded and unable to respond to service requests or be controlled by the attacker and refuse service requests~\cite{kaur2019distributed}. The service reliability of the server can be evaluated by calculating the ratio of the number of successful response requests $R_{s}^{suc}$ to the total number of requests $R_{s}^{tot}$ issued in the detection task. A higher ratio of successful responses indicates that the edge server $s$ has a stronger capability to process and complete service requests effectively. Therefore, the service performance of edge server $s$ can be calculated as
\begin{equation}
P_s^{ser}=\frac{R_s^{suc}}{R_s^{tot}}.
\label{eq11}
\end{equation}

The historical defense record of an edge server is an important indicator for evaluating its reputation value~\cite{wang2018secure}. The defense performance of an edge server can be determined by calculating the ratio of the number of successful defenses $N_{s}^{suc}$ to the total number of historical attacks $N_{s}^{tot}$, which can be calculated as $\beta_s=\frac{N_s^{suc}}{N_s^{tot}}$. Based on the above evaluation indicators, the reputation value of the edge server $s$ at the network communication layer can be calculated as~\cite{70}
\begin{equation}
Rep_s^{Net}=
\begin{cases}0,\quad \text{if $\beta_s \in (0, \theta_1)$},\\
\alpha\big(\omega P_s^{data}+(1-\omega)P_s^{ser}\big),\text{if $\beta_s \in (\theta_1, \theta_2)$},\\
\omega P_s^{data}+(1-\omega)P_s^{ser},\quad \text{otherwise}.
\end{cases}
\label{eq12}
\end{equation}

We divide \eqref{eq12} into three cases. In the first case, if the defense performance of edge server $s$ is lower than the threshold $\theta_{1}$, we consider it high-risk and set its reputation value of the network communication layer to 0, where $0<\theta_1<1$. In the second case, when the defense performance of the edge server $s$ is higher than the threshold $\theta_{1}$ but lower than the threshold $\theta_{2}$, we calculate the weighted sum of data security $P_s^{data}$ and service performance $P_s^{ser}$ of edge server $s$, adjusting the balance between the two with weight $\omega$ and penalizing with factor $\alpha$, where $\theta_{1}<\theta_{2}<1$ and $0<\alpha<1$. Finally, when the defense performance of the edge server is higher than the threshold $\theta_{2}$, we consider it to have a high degree of trust and directly use the weighted sum of data security and service performance to represent its reputation value.

The second step of trust evaluation occurs at the interaction layer between the VMUs and the edge server, reflecting the direct interaction experience between the VMUs and the edge server. The reputation value of the edge server is assessed by recording the evaluation given by the VMU during each interaction with the edge server. Let $\mathcal{E}_s=\{E_{1,s},\ldots, E_{i,s},\ldots, E_{n,s}\}$ denote the set of evaluations by all VMUs that have interacted with edge server $s$, where  $n$ denotes the number of VMUs interacting with server $s$. For each VMU $i$, the set $E_{i,s}=\{e_{i_{1},s},\ldots,e_{i_{m},s},\ldots,e_{i_{k},s}\}$ represents the evaluation scores across $k$ interactions, where $e_{i_{m},s}=1$ and $e_{i_{m},s}=0$ indicate a positive evaluation and a negative evaluation for the $m$-th interaction, respectively. In addition, the total number of evaluations received by edge server $s$ is calculated as $E_{s}^{tot}=\sum_{i=1}^{n} i_{k}$. Based on the beta distribution~\cite{35}, the reputation value of edge server $s$ at the interaction layer is defined as
\begin{equation}
Rep_s^{Int}=\frac{\Sigma_{i=1}^n\Sigma_{m=1}^ke_{i_m,s}+1}{E_s^{tot}+2}.
\label{eq13}
\end{equation}

In summary, when evaluating the reputation value of edge server $s$ using the two-layer trust evaluation model, the initial step involves calculating its $P_s^{data}$ and $P_s^{ser}$ at the network communication layer and then performing a weighted summation of $P_s^{data}$ and $P_s^{ser}$ to obtain the preliminary reputation value at this layer. The reputation value at the network communication layer is classified and discussed according to the historical defense data of edge server $s$. Then, the reputation value of edge server $s$ at the interaction layer is calculated based on the interaction evaluations from VMUs. Finally, the comprehensive reputation value of edge server $s$ is obtained by performing a weighted summation of the reputation values from the network communication layer and interaction layer, calculated as
\begin{equation}
Rep_s=\xi Rep_s^{Net}+(1-\xi) Rep_s^{Int},
\label{eq14}
\end{equation}
where $\xi$ is a weight parameter. In addition, the reputation value of the edge server $s$ is updated by combining the latest evaluated reputation value $Rep_s^{late}$ with the past reputation value $Rep_s^{past}$. The formula for updating is given by
\begin{equation}
Rep_s^{new}=\varphi Rep_s^{late}+(1-\varphi)Rep_s^{past},
\label{eq15}
\end{equation}
where $\phi \in (0,1)$ controls the update rate.

\subsection{Utility Function}
\label{s3-4}

To provide VMUs with a seamless and reliable service experience, we design the utility function of VMUs as the difference between reputation values and service latency. Specifically, the reputation value reflects the reliability and consistency of the edge server in providing services, while the latency measures the response and processing speed of the service, which directly affects the interactive experience of VMUs. High reputation values and low service latency lead to high utility value, indicating that VMUs can enjoy fast and reliable services. Therefore, the utility function of vehicle $v$ at time $t$ is given by
\begin{equation}
U_{v}(t)=\lambda(Rep_{s}(t)+Rep_{s_p}(t))-\mu T_{v}^{sum}(t).
\label{eq16}
\end{equation}

$Rep_s(t)$ and $Rep_{s_p}(t)$ are the reputation values of edge servers $s$ and $s_p$ selected by the vehicle $v$ at time $t$, respectively. $T_{v}^{sum}(t)$ is the total delay of the VT migration. $\lambda$ and $\mu$ are coefficients that convert unit reputation value and unit delay into monetary benefits. By applying the utility function, the experience benefit of the decision at a specific moment can be evaluated.

\section{Problem Formulation}
\label{s4}

This paper aims to maximize the overall utility of vehicles within a finite time $T$ by identifying the optimal migration decision strategy, subject to the constraints of edge server reliability and limited load, which is given by
\begin{subequations}
\begin{align}
\max_{K_v} & \sum_{t=1}^{T} \sum_{v=1}^{V} U_{v}(t) \label{eq17a}\\
\text{s.t.} \quad & k_{v,s}^{\text{sel}}(t) \in \{1,2, \ldots, S\}, \quad \forall v \in \mathcal{V}, \label{eq17b}\\
& k_{v,s_p}^{\text{sel}}(t) \in \{1,2, \ldots, S\}, \quad \forall v \in \mathcal{V}, \label{eq17c}\\
& K_{v}^{\text{pre}}(t) \in [0,1], \quad \forall v \in \mathcal{V}, \label{eq17d}\\
& P_{v,\mathfrak{s}}(t) \leq R_{\mathfrak{s}}^{\text{com}}, \quad \mathfrak{s}\in \{s, s_p\}, \forall v \in \mathcal{V},\label{eq17e}\\
& Rep_{\mathfrak{s}}(t) \geq Rep_{\mathfrak{s}}^{\text{thre}}, \quad \mathfrak{s} \in \{s, s_p\}, \label{eq17f}\\
& L_{\mathfrak{s}}(t) < L_{\mathfrak{s}}^{\text{max}}, \quad \mathfrak{s} \in \{s, s_p\}, \label{eq17g}\\
& t \in \{1, \ldots, T\}. \label{eq17h}
\end{align}
\end{subequations}

In constraints \eqref{eq17b} and \eqref{eq17c}, $k^{sel}_{v,s}$ and $k^{sel}_{v,s_p}$ represent the indices of the edge servers utilized by vehicle $v$ for VT migrations. This ensures that at any given time, the edge servers used for VT migrations by vehicles belong to the existing set of edge servers, thereby securing the feasibility of the migration decisions. Constraint \eqref{eq17d} specifies that the proportion of pre-migration tasks lies between $0$ and $1$. Constraint \eqref{eq17e} requires the communication ranges of the selected edge servers $s$ and $s_p$ to be greater than their Euclidean distances from the vehicle. Constraint \eqref{eq17f} demands that the reputation value of the selected server must exceed the preset threshold $Rep_{\mathfrak{s}}^{\text{thre}}$ to ensure the basic reliability of the task. Constraint \eqref{eq17g} ensures that the load on the chosen server at that moment does not reach its maximum capacity to prevent overload. Constraint \eqref{eq17h} indicates the need for optimization to be completed in a limited time frame $T$.

\section{Hybrid-GDM for Twin Migration}
\label{s5}

In this section, we first model the VT migration optimization problem as a POMDP and then propose the hybrid-GDM algorithm to generate the optimal migration decision.

\subsection{Problem Modeling}
\label{s5-1}

Since intelligent networks usually have the characteristics of high-dimensional configuration, non-linear relationships, and complex decision-making processes~\cite{65}, the optimal solution of intelligent networks typically changes with the dynamic environment~\cite{wen2024generative}. Therefore, complex network management models are required. GDMs can easily integrate environmental information into the denoising process~\cite{67} to capture high-dimensional and complex network structures. This avoids the problem that traditional DRL algorithms tend to converge to sub-optimal solutions in such tasks.

In the vehicular metaverse, there are multiple edge servers and vehicles. The objective is to complete the VT migration of all vehicles in a way that maximizes overall utility, i.e., \eqref{eq17a}. Given the limited communication distance and load of edge servers, the mobility pattern of each vehicle and the size of uploaded VT tasks are different, which makes POMDP particularly suitable for VT migrations. The unpredictability of future load on edge servers, as well as when and how potential attackers attack edge servers further justifies the POMDP-based approach. The POMDP model of the VT migration problem is described as follows:

\begin{itemize}
  \item [1)] \textbf{Observation Space:}
  The observation space $\mathcal{O}$ consists of real-time information from vehicles and edge servers. The observation space at time slot $t$ is defined as $\mathcal{O}(t)=\{P_v(t), T_v^{sum}(t),\forall v\in\mathcal{V}\}\cup\{L_s(t), Rep_s(t),\forall s\in\mathcal{S}\}$, where $P_v(t)$ is the position of vehicle $v$ at the current moment, $T_v^{sum}(t)$ is the total delay of a VT migration in time slot $t$, and $L_s(t)$ and $Rep_s(t)$ are the load condition and reputation value of edge server $s$ at this time, respectively.
  \item [2)] \textbf{Action Space:}
  The action space is defined as the migration decisions the agent generates for all vehicles, expressed as $\mathcal{A}=\{K_v,\forall v \in \mathcal{V} \}$. The migration decision of each vehicle at each time slot includes a discrete decision to select an edge server and a continuous decision to determine the proportion of pre-migration VT tasks. Thus, the migration decision of vehicle $v$ at time slot $t$ can be represented as the tuple $K_v(t) = \{K_v^{sel}(t), K_{v}^{pre}(t)\}$, where $K_v^{sel}(t) = \{k_{v,s}^{sel}(t), k_{v,s_p}^{sel}(t)\}$ represents the indices of the current edge server $s$ and the pre-migration edge server $s_p$ chosen by vehicle $v$ at time slot $t$. These indices correspond one-to-one with the edge servers in the vehicular metaverse. $K_{v}^{pre}(t) \in [0, 1]$ represents the proportion of the pre-migrated VT task in time slot $t$.
  \item [3)] \textbf{Reward Function:}
  In each time slot, the agent makes migration decisions $\mathcal{A}(t)$ for all vehicles based on the environment observation $\mathcal{O}(t)$. The environment returns a reward based on action $\mathcal{A}$. To maximize the overall utility, we define the reward as
  \begin{equation}
  \mathcal{R}(\mathcal{O}(t), \mathcal{A}(t))=\sum_{v=1}^{V} U_{v}(t).
  \label{eq18}
  \end{equation}
\end{itemize}

\subsection{Hybrid-GDM for Generating Optimal Decisions}
\label{s5-2}
\begin{figure*}[!h]
\centering
\includegraphics[width=0.9\textwidth]{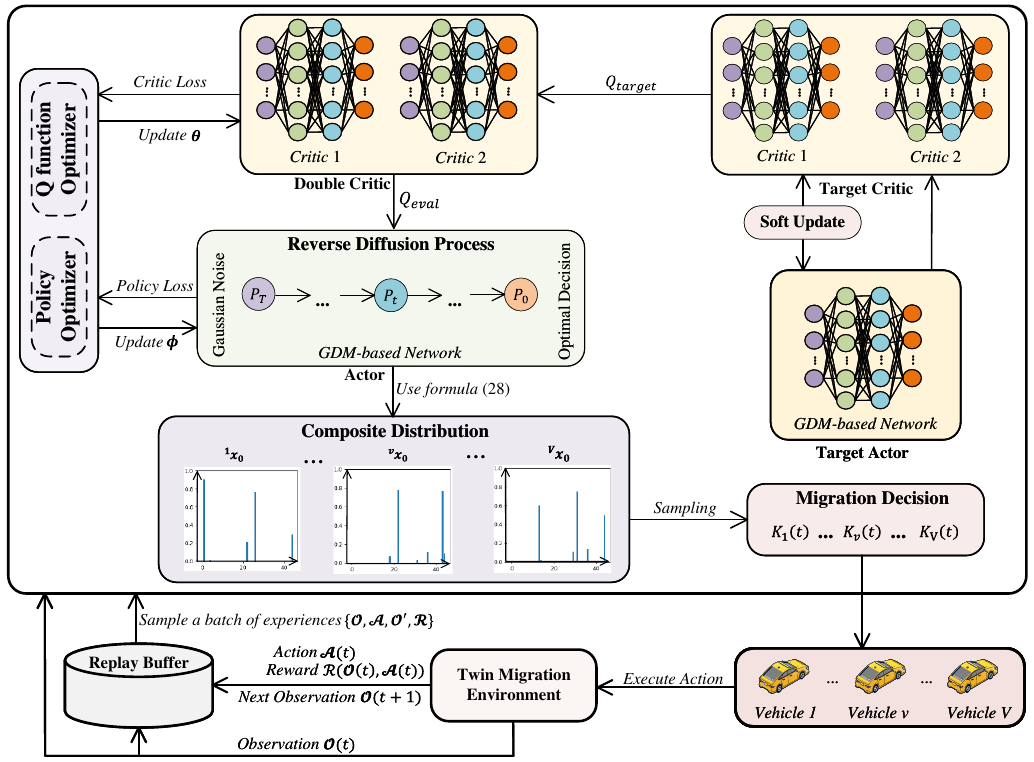}
\caption{The overall architecture of the hybrid-GDM algorithm.}
\label{fig3}
\end{figure*}

In this section, we propose the hybrid-GDM algorithm. The forward process of the diffusion model is first introduced, and then the backward process is shown how to generate optimal hybrid decisions from Gaussian noise. Finally, the overall architecture of the algorithm is proposed in Fig.~\ref{fig3}.
\subsubsection{Forward process of hybrid actions}
\label{s5-52-1}

VTs migration decision of vehicles is sampled from the composite distribution $\boldsymbol{x}_{0}=\pi_{\boldsymbol{\theta}}(\mathcal{O})\sim\mathbb{R}^{|\mathcal{A}|}$ of the output from the reverse denoising process based on the given observation $\mathcal{O}$. This distribution consists of three parts, i.e., $\boldsymbol{x}_0=\boldsymbol{x}_0^{s} \cup \boldsymbol{x}_0^{s_p} \cup \boldsymbol{x}_0^{pre}$, where $\boldsymbol{x}_0^{s}$ represents the probability of each edge server being chosen for uploading VT tasks, $\boldsymbol{x}_0^{s_p}$ represents the probability of each server being chosen as a pre-migration server, and $\boldsymbol{x}_0^{pre}$ represents the proportion of pre-migration tasks. We use $\boldsymbol{x}_t$ to represent the distribution of the forward process at step $t$, which is consistent in dimension with $\boldsymbol{x}_0$. This forward process starts from the target distribution $\boldsymbol{x}_0$ by adding a series of Gaussian noise, $\boldsymbol{x}_1,\ldots,\boldsymbol{x}_t,\ldots,\boldsymbol{x}_T$ is generated in sequence. In this process, the transformation from $\boldsymbol{x}_{t-1}$ to $\boldsymbol{x}_t$ can be defined as a normal distribution with $\sqrt{1-\beta_{t}}\boldsymbol{x}_{t-1}$ as the mean and $\beta_{t}\mathbf{I}$ as the variance determined by the variational posterior scheduler, given by~\cite{46}
\begin{equation}
q\left(\boldsymbol{x}_t|\boldsymbol{x}_{t-1}\right)=\mathcal{N}\left(\boldsymbol{x}_t;\sqrt{1-\beta_t}\boldsymbol{x}_{t-1},\beta_t\mathbf{I}\right).
\label{eq19}
\end{equation}

Since the system is modeled as a POMDP, $\boldsymbol{x}_t$ only depends on $\boldsymbol{x}_{t-1}$ from the previous step. The transitions $q\left(\boldsymbol{x}_t|\boldsymbol{x}_{t-1}\right)$ are multiplied over the denoising steps to form the distribution $\boldsymbol{x}_T$ conditioned on $\boldsymbol{x}_0$, denoted as~\cite{46}
\begin{equation}
q\left(\boldsymbol{x}_T|\boldsymbol{x}_0\right)=\prod_{t=1}^Tq\left(\boldsymbol{x}_t|\boldsymbol{x}_{t-1}\right).
\label{eq20}
\end{equation}

Based on~\eqref{eq20}, $x_t$ can be obtained through $t$ iterations of sampling. However, a large $t$ results in significant computational cost. To optimize the calculation method, by defining $\alpha_{t}=1-\beta_{t}$, $x_t$ can be obtained in a single sampling step, denoted as~\cite{du2024enhancing}
\begin{equation}
\mathbf{x}_t\sim q\left(\boldsymbol{x}_t\mid\boldsymbol{x}_0\right)=\mathcal{N}\left(\boldsymbol{x}_t;\sqrt{\bar{\alpha}_t}\boldsymbol{x}_0,\left(1-\bar{\alpha}_t\right)\mathbf{I}\right),
\label{eq21}
\end{equation}
where $\bar{\alpha}_{t}=\prod_{l=1}^{t}\alpha_{l}$ is the cumulative product of $\alpha_{l}$. 

In wireless networks, it is common for optimization problems to lack a data set for the optimal solutions~\cite{68} for the forward process, i.e., $\boldsymbol{x}_0$. Therefore, the forward process is not executed for the hybrid GDM, but the relationship between each parameter is constructed.

\subsubsection{Reverse process of hybrid actions}
\label{s5-52-2}

The optimal solutions $\boldsymbol{x}_0$ is inferred by progressively removing noise through the reverse process, beginning with a sample $\boldsymbol{x}_T$ drawn from a standard normal distribution  $\mathcal{N}(\mathbf{0}, \mathbf{I})$. However, estimating the conditional probability distribution $q\left(\boldsymbol{x}_{t-1}|\boldsymbol{x}_{t}\right)$ is challenging. Therefore, it is necessary to construct a parameterized model $p_{\boldsymbol{\theta}}$, denoted as
\begin{equation}
p_{\boldsymbol{\theta}}\left(\boldsymbol{x}_{t-1}|\boldsymbol{x}_{t}\right)=\mathcal{N}\left(\boldsymbol{x}_{t-1};\boldsymbol{\mu}_{\boldsymbol{\theta}}\left(\boldsymbol{x}_{t},t,\mathcal{O}\right),\tilde{\beta}_{t}\mathbf{I}\right),
\label{eq22}
\end{equation}
where $\tilde{\beta}_{t}=\frac{1-\bar{\alpha}_{t-1}}{1-\bar{\alpha}_{t}}\beta_{t}$ denotes the deterministic variance amplitude which can be computed, and the mean $\boldsymbol{\mu}_{\boldsymbol{\theta}}$ is predicted through a deep neural network~\cite{46}.

According to \eqref{eq21}, by applying hyperbolic tangent activation, the generated noise is reduced to avoid high levels of noise that obscure the actual action distribution. The obtained reconstructed sample $\boldsymbol{x}_{0}$ is
\begin{equation}
\boldsymbol{x}_0=\frac{1}{\sqrt{\bar{\alpha}_t}}\boldsymbol{x}_t-\sqrt{\frac{1}{\bar{\alpha}_t}-1}\cdot\tanh\left(\boldsymbol{\epsilon}_{\boldsymbol{\theta}}(\boldsymbol{x}_t,t,\mathcal{O})\right),
\label{eq23}
\end{equation}
where $\boldsymbol{\epsilon}_{\boldsymbol{\theta}}(\boldsymbol{x}_t,t,\mathcal{O})$ a neural network parameterized to generate denoising noise based on observation $\mathcal{O}$.

Calculating $\boldsymbol{x}_0$ by using \eqref{eq23} is unfeasible due to the emergence of additional noise in each reverse denoising step. This noise source is distinct from the noise introduced during the forward process. Thus,  we employ the Bayesian formula to reformulate the reverse computation into forward ones, thereby converting it into a Gaussian probability distribution. The relationship between the mean $\boldsymbol{\mu}_{\boldsymbol{\theta}}$ and $\boldsymbol{x}_0$ is
\begin{equation}
\boldsymbol{\mu}_{\boldsymbol{\theta}}\left(\boldsymbol{x}_t,t,\mathcal{O}\right)=\frac{\sqrt{\alpha_t}\left(1-\bar{\alpha}_{t-1}\right)}{1-\bar{\alpha}_t}\boldsymbol{x}_t+\frac{\sqrt{\bar{\alpha}_{t-1}}\beta_t}{1-\bar{\alpha}_t}\boldsymbol{x}_0,
\label{eq24}
\end{equation}
where $t=1,\ldots, T$. Based on \eqref{eq23} and \eqref{eq24}, the estimated mean is rewritten as
\begin{equation}
\boldsymbol{\mu}_{\boldsymbol{\theta}}\left(\boldsymbol{x}_t,t,\mathcal{O}\right)=\frac{1}{\sqrt{\alpha_t}}\left(\boldsymbol{x}_t-\frac{\beta_t\tanh\left(\boldsymbol{\epsilon}_{\boldsymbol{\theta}}(\boldsymbol{x}_t,t,\mathcal{O})\right)}{\sqrt{1-\bar{\alpha}_t}}\right).
\label{eq25}
\end{equation}

Then, we can start from the standard Gaussian distribution $p\left(\boldsymbol{x}_{T}\right)$, and gradually substitute $t=T,T-1,\ldots,1$ in transition distribution $p(\boldsymbol{x}_t)p_\theta(\boldsymbol{x}_{t-1}|\boldsymbol{x}_t)$ to obtain $p_{\boldsymbol{\theta}}\left(\boldsymbol{x}_{0}\right)$, given by
\begin{equation}
p_{\boldsymbol{\theta}}\left(\boldsymbol{x}_{0}\right)=p\left(\boldsymbol{x}_{T}\right)\prod_{t=1}^{T}p_{\boldsymbol{\theta}}\left(\boldsymbol{x}_{t-1}|\boldsymbol{x}_{t}\right).
\label{eq26}
\end{equation}

To back-propagate gradients from random variables sampled from the distribution, the authors in~\cite{20} adopted reparameterization to separate the randomness from the parameters of the distribution. This allows the gradient to be back-propagated through the new parameterized path. The updated rules are given by
\begin{equation}
\boldsymbol{x}_{t-1}=\boldsymbol{\mu}_{\boldsymbol{\theta}}\left(\boldsymbol{x}_{t},t,\mathcal{O}\right)+\left(\tilde{\beta}_{t}/2\right)^{2}\odot\boldsymbol{\epsilon}.
\label{eq27}
\end{equation}

Starting from randomly generated normal noise, $\boldsymbol{x}_{t} (0\leq t<T)$ and the final output $\boldsymbol{x}_{0}$ can be obtained by iterating the backward update rule \eqref{eq27}.

Then, we extract and individually process the composite distribution $^{v}\boldsymbol{x}_{0}$ corresponding to each vehicle from $\boldsymbol{x}_{0}=\{^{v}\boldsymbol{x}_{0}, \forall v \in \mathcal{V}\}$. For the composite distribution $^v\boldsymbol{x}_{0}$ of vehicle $v$, we apply the softmax function $f(\boldsymbol{z}_i)=\frac{e^{\boldsymbol{z}_i}}{\sum_{j=1}^Se^{\boldsymbol{z}_j}}(i=1,2,..., S)$ to process $^{v}\boldsymbol{x}_{0}^{s}$ and $^{v}\boldsymbol{x}_{0}^{s_p}$, respectively, convert them into the probability distributions for choosing edge servers $s$ and $s_p$, and keep the proportion of pre-migration tasks $^v\boldsymbol{x}_{0}^{pre}$ as the original value. The processed distribution $\boldsymbol{x}_0'$ is
\begin{equation}
\boldsymbol{x}_0'=\{f(^v\mathbf{x}_0^{s})\cup f(^v\mathbf{x}_0^{s_p})\cup ^v\mathbf{x}_0^{pre},\forall v \in \mathcal{V}\}.
\label{eq28}
\end{equation}

To enhance the exploration of strategies during the training phase, minor noise is added to the proportion $\{^v\mathbf{x}_0^{pre}, \forall v \in \mathcal{V}\}$ of pre-migration tasks for all vehicles, and indices $\{K_{v}^{sel},\forall v \in \mathcal{V}\}$ of edge servers $s$ and $s_p$ are sampled from the probability distributions $\{^v\boldsymbol{x}_{0}^{s},\forall v \in \mathcal{V}\}$ and $\{^{v}\boldsymbol{x}_{0}^{s_p},\forall v \in \mathcal{V}\}$, respectively. During the evaluation phase, the proportion $\{^v\mathbf{x}_0^{pre},\forall v \in \mathcal{V}\}$ of the pre-migration task is maintained at its original value. For $\{K_{v}^{sel},\forall v \in \mathcal{V}\}$, the edge servers $s$ and $s_p$ with the highest probabilities from the corresponding distributions are selected.

\subsubsection{Algorithm architecture}
\label{s5-2-3}

The overall framework of the hybrid-GDM algorithm is shown in Fig.~\ref{fig3}, which contains several components that work together to promote strategy optimization jointly. These components include actor, double critic, target networks, experience replay buffer, and twins migration environment. 

In each environment step $t$, the agent receives observations $\mathcal{O}(t)$ and uses the diffusion model to output hybrid action distribution $\pi_{\theta}(\mathcal{O}({t}))$, where $\theta$ are the parameters of the GDM-based network. Then, actions $\mathcal{A}(t)=\{K_{v}(t),\forall v \in V)\}$ are sampled from the action distribution and fed back to the environment. The environment transitions to state $\mathcal{O}({t+1})$, and rewards $\mathcal{R}(\mathcal{O}(t), \mathcal{A}(t))$ are returned to agent. Finally, the agent records the experience ($\mathcal{O}(t), \mathcal{A}(t), \mathcal{O}({t+1}), \mathcal{R}(\mathcal{O}(t), \mathcal{A}(t))$) into the experience replay buffer. These steps will be executed $K$ times before the agent learns the policy.

Actor-network $\pi_{\theta}$ extracts a minimum batch of data from the experience replay buffer for optimization during the policy improvement. To reduce overestimation bias, a double critic network is utilized to implement the Q function ${Q}_{\phi}(\mathcal{O}(t),\mathcal{A}(t))$. Unlike the standard Q function~\cite{liu2021drl}, this Q function generates a vector of Q values $\boldsymbol{q}\in\mathbb{R}^{|\mathcal{A}|}$ with the same dimensions as the action distribution to evaluate the value of discrete and continuous actions comprehensively. Each critic network has its own parameters, i.e., ${\boldsymbol{\phi}}=\{\boldsymbol{\phi}^1,\boldsymbol{\phi}^2\}$, and is updated separately with an identical optimization objective. In the training phase, the minimum Q value estimate among the two critic networks is chosen to update the actor-network, i.e., $\boldsymbol{q}=Q_{\boldsymbol{\phi}}\left(\mathcal{O}(t),\mathcal{A}(t)\right)=\min\left\{Q_{\boldsymbol{\phi}^{1}}\left(\mathcal{O}(t),\mathcal{A}(t)\right), Q_{\boldsymbol{\phi}^{2}}\left(\mathcal{O}(t),\mathcal{A}(t)\right)\right\}$.

The Q value estimates the expected cumulative reward of each action within the given observation. To update the actor network toward selecting actions with higher Q values, we transform the optimization objective into a minimization problem as
\begin{equation}
\min_{\boldsymbol{\theta}}\left\{-\pi_{\boldsymbol{\theta}}\left(\mathcal{O}(t)\right)^{T}Q_{\boldsymbol{\phi}}\left(\mathcal{O}(t),\mathcal{A}(t)\right)\right\}.
\label{eq29}
\end{equation}

We use the Adam gradient descent algorithm to address the optimization problem. The gradient corresponding to~\eqref{eq29} is computed by sampling experiences of batch size $\mathcal{B}$ at $e$-th training iteration and is calculated as
\begin{equation}
\mathbb{E}_{\mathcal{O}(t)\sim\mathcal{B}_e}\begin{bmatrix}-\nabla_{\boldsymbol{\theta}}\boldsymbol{\phi}_{\boldsymbol{\theta}}(\mathcal{O}(t))^TQ_\phi(\mathcal{O}(t),\mathcal{A}(t))\end{bmatrix}.
\label{eq30}
\end{equation}
The agent learns the optimal policy parameters by iteratively executing the gradient update formula, with the learning rate $\eta_{\mathrm{a}}$ controlling the step size of parameter adjustments. The gradient update formula is
\begin{align}
\boldsymbol{\theta}_{e+1} \leftarrow & \ \boldsymbol{\theta}_e - \eta_{\mathrm{a}} \cdot \left( \mathbb{E}_{\mathcal{O}(t) \sim \mathcal{B}_e} \left[ -\nabla_{\boldsymbol{\theta}} \pi_{\boldsymbol{\theta}}\left(\mathcal{O}(t)\right)^T \times \right. \right. \nonumber \\
& \left. \left. Q_{\boldsymbol{\phi}}\left(\mathcal{O}(t), \mathcal{A}(t)\right) \right] \right).
\label{eq31}
\end{align}

Only when the critic network's estimate of the expected cumulative reward is sufficiently accurate can the actor network find the optimal policy. Therefore, we update the critic network by reducing the temporal difference error between the Q target and Q evaluation, denoted as
\begin{equation}
\begin{aligned}
&\min_{\phi^{1},\phi^{2}} && \mathbb{E}_{(\mathcal{O}(t), \mathcal{A}(t), \mathcal{O}(t+1), \mathcal{R}(\mathcal{O}(t), \mathcal{A}(t))) \sim \mathcal{B}_{e}} \left[\sum_{i=1}^2\left(\hat{y}_{e} - y_{e}^{i}\right)^{2}\right] \\
&\text{s.t.} && y_{e}^{i} = Q_{\boldsymbol{\phi}_{e}^{i}}(\mathcal{O}(t), \mathcal{A}(t)), \\
&&& \hat{y}_{e} = \mathcal{R}(\mathcal{O}(t), \mathcal{A}(t)) + \gamma(1 -f_{t+1}) \hat{\pi}_{\hat{\boldsymbol{\theta}}_{e}} \cdot \\
&&& \qquad (\mathcal{O}(t+1))^{T}\hat{Q}_{\hat{\boldsymbol{\phi}}_{e}}(\mathcal{O}(t+1), \mathcal{A}(t+1)),
\end{aligned}
\label{eq32}
\end{equation}
where $\gamma$ is the discount factor for future rewards, and $f_{t+1}$ is a binary variable indicating the termination flag.

We use target networks to improve training stability by freezing their parameters during gradient descent and updating them slowly via a soft update mechanism, denoted as
\begin{equation}
\begin{aligned}
\hat{\boldsymbol{\theta}}_{e+1} &\leftarrow \tau\boldsymbol{\theta}_{e} + (1-\tau)\hat{\boldsymbol{\theta}}_{e}, \\
\hat{\boldsymbol{\phi}}_{e+1} &\leftarrow \tau\boldsymbol{\phi}_{e} + (1-\tau)\hat{\boldsymbol{\phi}}_{e},
\end{aligned}
\label{eq33}
\end{equation}
where $\hat{\boldsymbol{\theta}}_{e}$ and $\hat{\boldsymbol{\phi}}_{e}$ are the parameters of the target actor network and target critic network, respectively, and $\tau\in(0,1]$ controls the frequency at which the target network is updated. Finally, the policy update and Q function update are iteratively executed until convergence, thereby maximizing the utility objective \eqref{eq17a}. The detailed algorithm of hybrid-GDM is shown in Algorithm~\ref{algorithm1}.

The computational complexity of Algorithm~\ref{algorithm1} depends on experience collection and parameter updating. Specifically, the computational complexity of experience collection is $O(EK\left(V+T|\boldsymbol{\theta}|\right))$. This includes the overhead of interacting with the environment $O(EKV)$, where $E$ is the number of training epochs, $K$ is the number of experiences collected, and $V$ is the computational cost of a single step when interacting with the environment. Additionally, since the reverse diffusion process performs $T$ steps of denoising, the additional overhead generated is $O(T|\boldsymbol{\theta}|)$, where $|\boldsymbol{\theta}|$ is the number of parameters in the actor-network. The computational complexity of parameter update is $O(E\left(b+1\right)(\left|\boldsymbol{\theta}\right|+\left|\boldsymbol{ \phi}\right|))$. This includes the overhead $O\left(Eb\left(|\boldsymbol{\theta}|+|\boldsymbol{\phi}|\right)\right)$ of updating both the actor-network and the critic-network as well as the overhead of updating the target networks $O\left (E\left(|\boldsymbol{\theta}|+|\boldsymbol{\phi}|\right)\right)$, where $b$ is the batch size, and $|\boldsymbol{\phi}|$ is the number of parameters in the critic-network. Therefore, the computational complexity of Algorithm~\ref{algorithm1} is $O\left(E\left[KV+TK|\boldsymbol{\theta}|+(b+1)\left(|\boldsymbol{\theta}|+|\boldsymbol{ \phi}|\right)\right]\right)$.

\begin{algorithm}[!t]
\caption{Hybrid-GDM Algorithm}
\begin{algorithmic}[1]
\STATE Initialize actor-network parameters $\boldsymbol{\theta}$, critic-network parameters $\boldsymbol{\phi}$, target-network parameters $\hat{\boldsymbol{\theta}} \gets \boldsymbol{\theta}$, $\hat{\boldsymbol{\phi}} \gets \boldsymbol{\phi}$, observation $\mathcal{O}$, and experience replay buffer $\mathcal{D}$;
\FOR{the training epoch $e$ = 1 to $E$}
    \FOR{the collected experiences $k$ = 1 to $K$}
        \STATE Obtain the observation $\mathcal{O}$ and initialize the standard normal distribution $\boldsymbol{x}_{T}\sim\mathcal{N}(\boldsymbol{0},\mathbf{I})$.
        \FOR{the denoising step $t$ = $T$ to 1}
            \STATE Use a deep neural network to infer and scale a denoising distribution $\mathrm{tanh}\left(\boldsymbol{\epsilon}_{\boldsymbol{\theta}}(\boldsymbol{x}_{t},t,\mathcal{O})\right)$;
            \STATE Use~\eqref{eq25} to compute the mean of the parameterized model $p_{\boldsymbol{\theta}}\left(\boldsymbol{x}_{t-1}|\boldsymbol{x}_{t}\right)$;
            \STATE Use the reparameterization method~\eqref{eq27} to compute the composite distribution $\boldsymbol{x}_{t-1}$;
        \ENDFOR
        \STATE Process the composite distribution of $\boldsymbol{x}_{0}$ with~\eqref{eq28};
        \STATE Add the exploration noise to $\boldsymbol{x}_{0}$, and then sample the hybrid actions $\mathcal{A}$ from this distribution;
        \STATE Perform action $\mathcal{A}$ within the environment and obtain the observation $\mathcal{O}^{'}$ and reward $\mathcal{R}$;
        \STATE Record experience tuple $(\mathcal{O}, \mathcal{A}, \mathcal{O}', \mathcal{R})$ in replay buffer $\mathcal{D}$;
    \ENDFOR
    \STATE Extract a batch of experiences $\mathcal{B}=(\mathcal{O},\mathcal{A},\mathcal{O}^{'},\mathcal{R})$ from the experience replay buffer $\mathcal{D}$ for optimization;
    \STATE Update the actor-network $\boldsymbol{\theta}$ according to \eqref{eq31};
    \STATE Update the critic network $\boldsymbol{\phi}$ with one gradient descent step to minimize \eqref{eq32};
    \STATE Perform a soft update of the target network parameters $\hat{\boldsymbol{\theta}}$, $\hat{\boldsymbol{\phi}}$ using \eqref{eq33};
\ENDFOR
\RETURN the optimal migration decision solution $\mathcal{A}^0$.
\end{algorithmic}
\label{algorithm1}
\end{algorithm}

\section{Numerical Results}
\label{s6}

In this section, we initially show the experimental scenario configurations. We then analyze the convergence of the hybrid-GDM algorithm and other baselines. Finally, we comprehensively evaluate the performance of the hybrid-GDM algorithms in comparison to other baseline algorithms across multiple scenario configurations.

\subsection{Parameter Settings}
\label{s6-1}

In our scenario, there are 22 edge servers, including 20 RSUs and 2 satellites designed to supplement computational capabilities. This setup aims to ensure a seamless experience for VMUs even when the edge servers are unevenly deployed and under heavy load. All RSUs communicate with each other via wired networks, while the connection with satellites is established through wireless networks. Besides, we consider attackers in the scenario who irregularly launch attacks on edge servers, including direct DDoS attacks, indirect DDoS attacks, and co-resident attacks. All edge servers are at risk of being compromised. Moreover, each vehicle has a unique driving trajectory. The essential parameters for the experiment are presented in Table ~\ref{tab1}.

\begin{table}[!t]
\centering
\caption{Key Parameter Setting}
\label{tab1}
\begin{tabular}{>{\centering\arraybackslash}p{1.1cm}>{\raggedright\arraybackslash}p{3.85cm}>{\centering\arraybackslash}p{1.95cm}} \toprule
 \textbf{Parameters}& \textbf{Descriptions}& \textbf{Values}\\ \midrule
    $V$&Number of vehicles & $10$ \\ 
    $S$&Number of edge servers & $22$ \\ 
    $c_{\mathcal{S}}$&Computing capability range of all edge servers&$\text{[$100$, $200$] MHz}$~\cite{14} \\
    $D_{\mathcal{V}}^{up}$&Size range of VT tasks uploaded by all vehicles& $\text{[$15$,$175$] MB}$~\cite{14}\\  
    $B_{\mathcal{S}}$&Migration bandwidth range for all edge servers&$\text{[$500$, $900$] Mbps}$~\cite{14} \\
    $\xi$& Weight coefficient between the reputation values of the network communication layer and the interaction layer&$0.5$\\
    $\lambda$&Coefficient of monetary benefit per unit reputation value &$4$\\
    $\mu$ & Coefficient of monetary gain per unit delay&$1$\\ 
    $E$&Number of episodes & $10^{5}$\\
    $T$&  Denoising steps&$5$\\
    $D$&  Maximum capacity of the replay buffer&$1\times10^{6}$\\
     $K$&  Number of transitions per training step&$1\times10^{3}$\\
    $\eta_{\mathrm{a}}$&  Learning rate of the actor-networks&$1\times10^{-4}$~\cite{20}\\
    $\eta_{\mathrm{c}}$& Learning rate of the critic networks&$1\times10^{-3}$~\cite{20}\\
    $\tau$& Weight of soft update&$0.005$~\cite{20}\\
    $\gamma$& Discount factor &$0.95$\\\bottomrule
\end{tabular}
\end{table}

\subsection{Convergence Analysis}
\label{s6-2}

\begin{figure}[!t]
\centering
\includegraphics[width=0.45\textwidth]{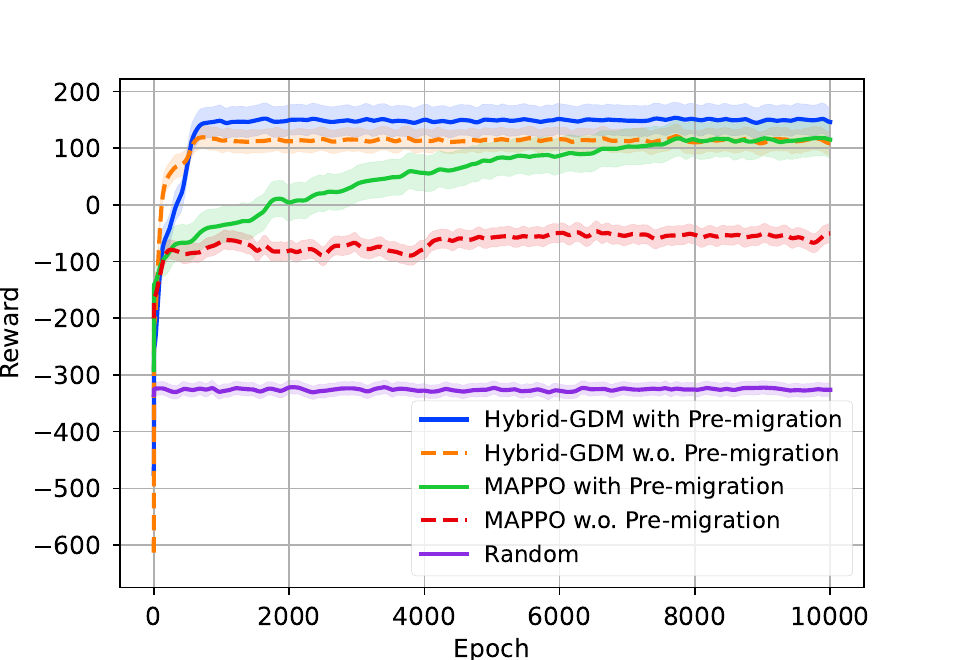}
\caption{Comparison of test reward curves of different algorithms.}
\label{fig4}
\end{figure}

As shown in Fig.~\ref{fig4}, we compare the hybrid-GDM algorithm with several baselines to verify our algorithm's effectiveness. These baseline algorithms include Hybrid-GDM w.o. pre-migration, MAPPO with pre-migration, MAPPO w.o. pre-migration, and random strategies, where 
w.o. pre-migration means that VT tasks are not considered pre-migrated when training using the relevant algorithm, i.e., these tasks are always processed on the current edge server $s$ without being pre-migrated to another edge server $s_p$.

Through the reward curve, we observe that the hybrid-GDM outperforms these baseline algorithms in both convergence speed and convergence value. Compared with Hybrid-GDM w.o. pre-migration, MAPPO with pre-migration, and MAPPO w.o. pre-migration, the performance is improved by 6.06$\%$, 25.04$\%$, and 62.52$\%$, respectively. This is because GDMs can incorporate dynamically changing environmental information (e.g., the load on edge servers, the location of vehicles, and the reputation values of edge servers) into the denoising process. After sufficient training, GDMs can generate the optimal VT migration strategy under any environmental conditions. In addition, the algorithm implementing pre-migration shows a higher convergence value, which shows that pre-migration can effectively reduce the load pressure of the current edge server $s$ and transfer it to the pre-migration server $s_p$. Especially for vehicles that move faster and stay within the current edge server coverage for a shorter time, it is necessary to pre-migrate more tasks to the edge server $s_p$. It is worth noting that even if the hybrid-GDM algorithm does not consider pre-migration, its final convergence value is almost the same as that of the MAPPO algorithm that considers pre-migration, and the convergence speed is faster, which highlights the advantages of hybrid-GDM in decision generation.

\subsection{Performance Evaluation}
\label{s6-3}

\begin{figure}[!t]
\centering
\includegraphics[width=0.45\textwidth]{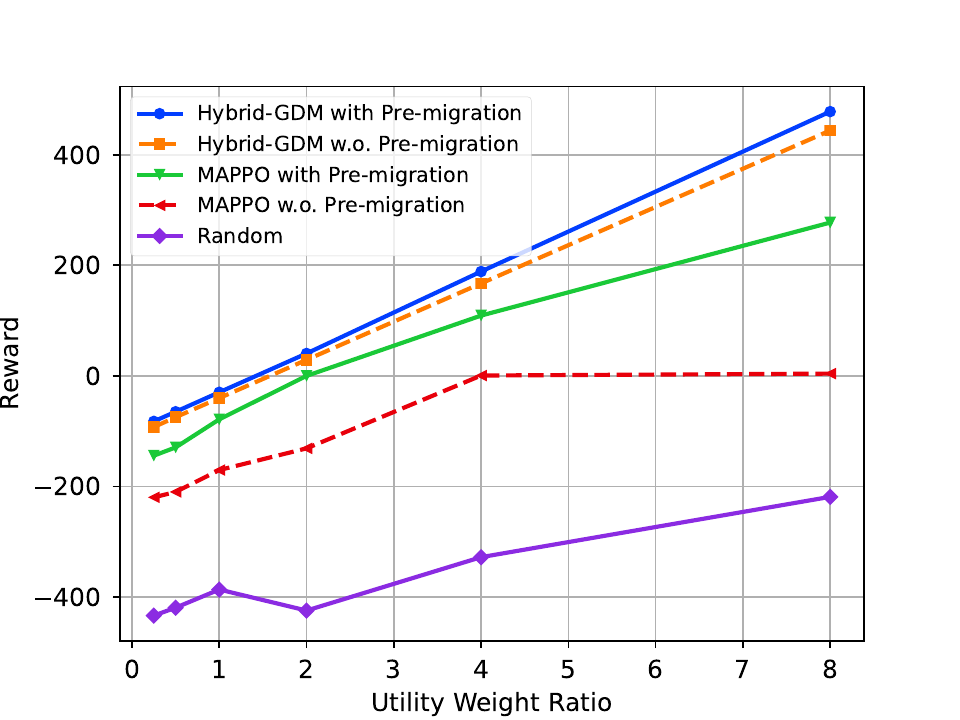}
\caption{Rewards under different utility weight ratios.}
\label{fig5}
\end{figure}

To evaluate the robustness of the hybrid-GDM algorithm under different system configurations, we compared the performance of various algorithms under different system parameters. In the utility function \eqref{eq16}, $\lambda$ and $\mu$ are coefficients that convert unit reputation value and unit delay to experience benefits. As these coefficients may vary across application scenarios, we introduce the utility weight ratio $\rho$ to represent the ratio between $\lambda$ and $\mu$, i.e., $\rho=\frac{\lambda}{\mu}$. Specifically, we consider several different values of $\rho$ (i.e., ${0.25, 0.5, 1, 2, 4, 8}$) to evaluate the rewards of agents under different settings of $\rho$, as shown in Fig. \ref{fig5}. The proposed hybrid-GDM algorithm demonstrates optimal performance across all $\rho$ settings, with performance improvements of 4.43$\%$, 22.4$\%$, and 56.86$\%$ compared to the baseline Hybrid-GDM w.o. pre-migration, MAPPO with pre-migration, and MAPPO w.o. pre-migration, respectively. As the proportion of reputation value in the total reward increases, the reward difference between the scheme that does not use the pre-migration and the scheme that uses the pre-migration gradually expands, especially in the MAPPO algorithm. The reason is that adopting a pre-migration decision can increase the available migration options, when an attacker launches an attack, effectively mitigating the impact of attacks on edge servers by choosing pre-migration edge servers with higher reputation values and altering the proportion of pre-migration tasks, thus enhancing VMU's utility.

\begin{figure}[!t]
\centering
\includegraphics[width=0.45\textwidth]{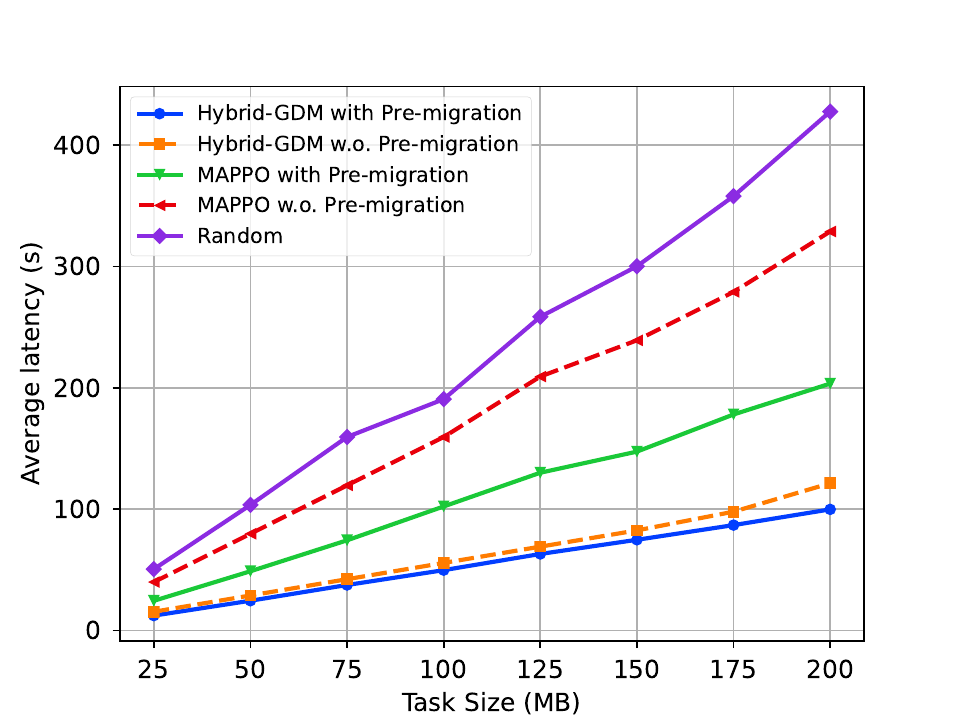}
\caption{Average system latency under different task sizes.}
\label{fig6}
\end{figure}

Figure~\ref{fig6} shows the changes in average system latency across eight levels of VT task sizes, ranging from 25 MB to 200 MB. The average system latency also rises as the size of the VT tasks increases, but the hybrid-GDM algorithm always has the lowest system delay, which reduces the average system delay by 3.9$\%$, 24.5$\%$, and 54.45$\%$ respectively compared to the baseline Hybrid-GDM w.o. Pre-migration, MAPPO with Pre-migration and MAPPO w.o. Pre-migration. It can be seen that when the VT task size is small, an edge server can handle the load without a significant impact on system latency, regardless of the use of pre-migration decisions. Thus, the latency difference between schemes employing pre-migration and those not using them remains small. However, as VT task sizes increase, this gap widens. Pre-migration helps distribute the load from the current edge server to the pre-migrated edge server, reducing the additional queuing delay caused by not using the pre-migration strategy.

\begin{figure}[!t]
\centering
\includegraphics[width=0.45\textwidth]{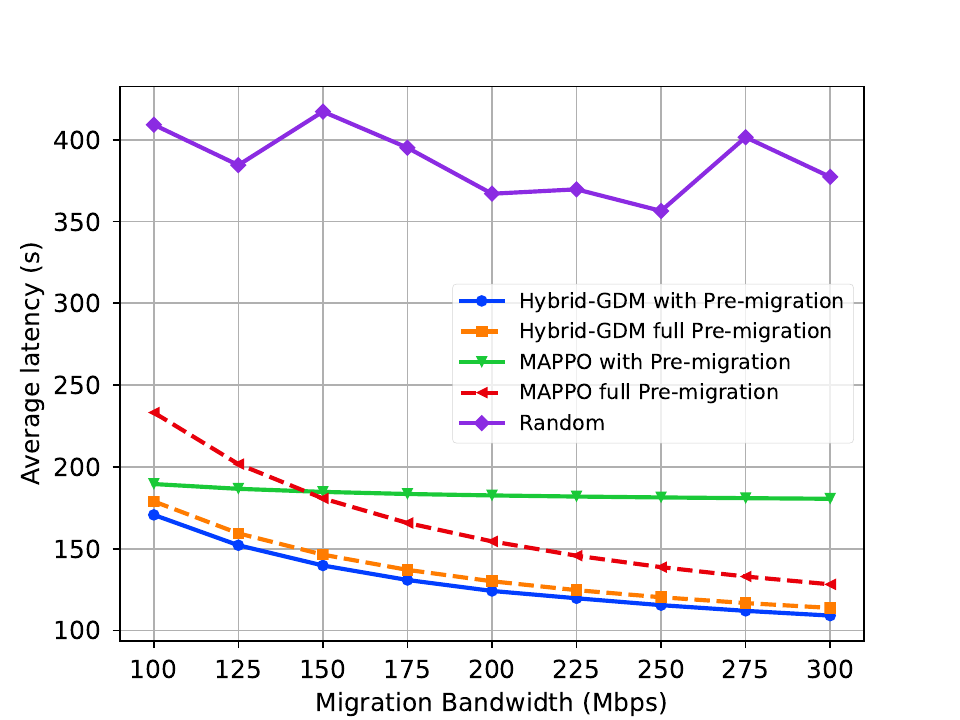}
\caption{Average system latency under different migration bandwidths.}
\label{fig7}
\end{figure}

Figure~\ref{fig7} shows the changes in average system latency across nine levels of migration bandwidth. The hybrid-GDM algorithm achieves a reduction in average system latency compared to the baseline Hybrid-GDM full Pre-migration, MAPPO with Pre-migration, and MAPPO full Pre-migration by 1.54 $\%$, 13.73 $\%$, and 8.85 $\%$, respectively. Note that full pre-migration denotes that the pre-migration proportion of VT tasks is set to 1. It can be observed that as the migration bandwidth increases, the average latency of MAPPO full Pre-migration initially exceeds but eventually falls below that of MAPPO with Pre-migration. The reason is that both the current edge server $s$ and the pre-migration edge server $s_p$ simultaneously process the unfinished tasks. With full pre-migration of VT tasks, higher migration bandwidths allow the substitution of queuing delays at the current server with lower migration delays, thereby reducing the total system latency. Owing to the effective scheme generation capabilities of the hybrid-GDM, it still exhibits lower latency than Hybrid-GDM full Pre-migration.

\begin{figure}[!t]
\centering
\includegraphics[width=0.45\textwidth]{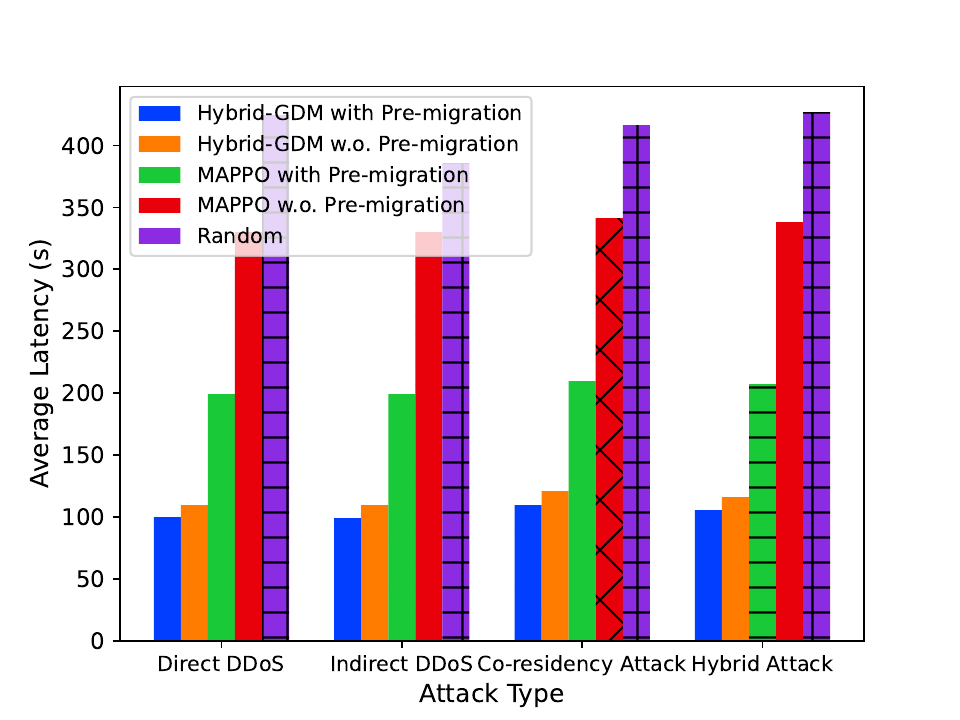}
\caption{Average latency under different attack types.}
\label{fig8}
\end{figure}
\begin{figure}[!t]
\centering
\includegraphics[width=0.45\textwidth]{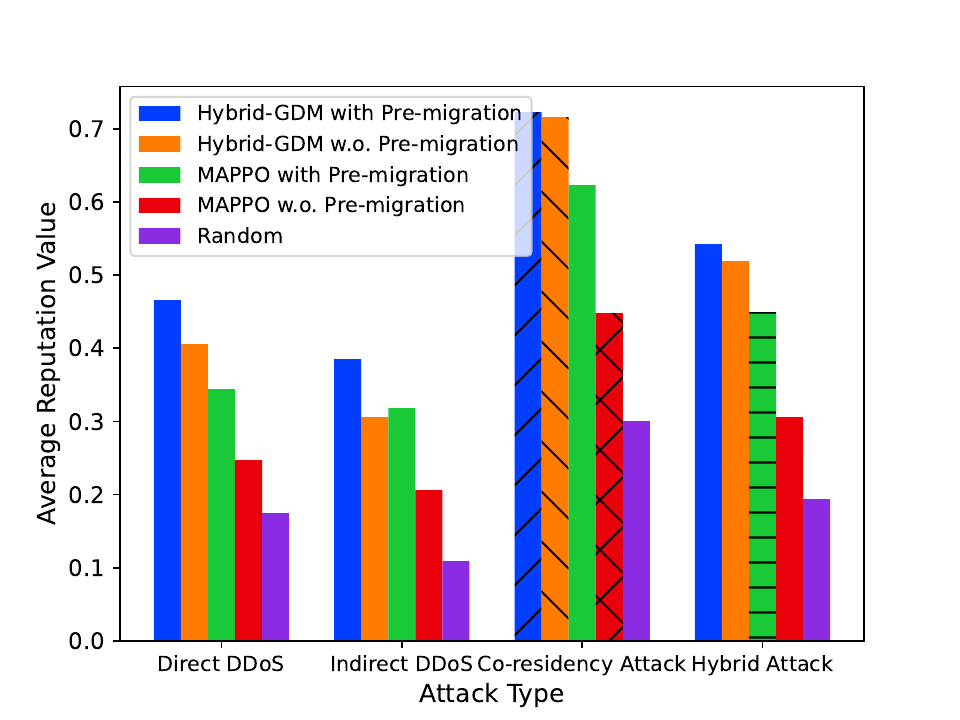}
\caption{Average reputation value of edge servers selected by vehicle under different attack types.}
\label{fig9}
\end{figure}

Figures~\ref{fig8} and~\ref{fig9} show the changes in average system latency and the mean reputation value of selected edge servers when subjected to different types of attacks. Four attack scenarios were considered, i.e., direct DDoS attacks only, indirect DDoS attacks only, co-residence attacks only, and a hybrid attack of the first three attacks. The hybrid-GDM algorithm consistently demonstrated the lowest average system latency and the highest mean reputation scores in all four scenarios. We observe that due to the concealment of co-residence attacks, the average reputation value of edge servers is relatively high in the co-residence attacks scenario, but the impact of co-residence attacks on latency is still reflected in Fig.~\ref{fig8}. It proves that our solution can effectively reduce system latency in different scenarios, thereby improving the utility of VMUs.

\section{Conclusion}
\label{CONCLUSION}

In this paper, we studied the VT migration problem in vehicular metaverses. To ensure a seamless virtual experience for VMUs, we proposed a secure and reliable VT migration framework. Considering the potential security threats of edge servers, we designed a two-layer trust evaluation model and modeled the mainstream attack methods in VT migration scenarios. The VT migration problem is formulated as POMDP, and we proposed a hybrid-GDM algorithm that can perform discrete actions and continuous actions simultaneously to ensure that the utility is maximized. Numerical results demonstrate that the proposed scheme can achieve low latency for VT migration while ensuring that the vehicle selects edge servers with high reputation values. For future work, we will apply the hybrid-GDM algorithm to more attacks and focus on detecting and identifying abnormal edge servers in vehicular metaverses.

\bibliographystyle{IEEEtran}
\bibliography{ref}

\end{document}